\documentclass{article}
\usepackage{spconf,amsmath,graphicx}

\newcommand{\CTCTWO}{CTC${}^2$}

\title{Advances in All-Neural Speech Recognition}

\name{Geoffrey Zweig, Chengzhu Yu, Jasha Droppo and Andreas Stolcke}
\address{Microsoft Research}

\begin{document}
\ninept

\maketitle

\begin{abstract}
This paper advances the design of CTC-based all-neural (or end-to-end)
speech recognizers. We propose a novel symbol inventory,
and a novel
iterated-CTC method in which a second system is used to transform a noisy
initial output into a cleaner version. 
We present a number of stabilization and initialization methods we
have found useful in training these networks.

We evaluate our system on the commonly
used NIST 2000 conversational telephony test set, and significantly 
exceed the previously published performance of similar systems,
both with and without the use of an external language model and decoding
technology. 
\end{abstract}


\section{Introduction}
\label{sec:intro}

In the recent renaissance of neural network speech recognition
\cite{dahl2011large,mohamed2009deep,seide2011conversational,sak2014long,hinton2012deep}, 
as well as in pioneering earlier work \cite{bourlard2012connectionist,robinson1991recurrent}, the
networks have been mainly used as a drop-in replacement
for the acoustic model in an HMM system.
They have also been used for feature-augmentation in 
a ``tandem'' GMM-HMM system 
\cite{hermansky2000tandem}, which again relied on a standard HMM backbone.
The state-of-the-art today is a hybrid HMM/neural-net
system, which uses the classical decoding strategy
\begin{equation} 
w^\ast = \arg\max_w P(w) P(a|w) 
\label{eq:dcd}
\end{equation}
where the prior on words $P(w)$ is estimated with a language model trained on 
text only, and the probability of the acoustics $P(a|w)$ is estimated with a 
neural network acoustic model. The acoustic model still uses a decision 
tree \cite{young1994tree} to further decompose the word sequence into context-dependent
triphone states, and a decoder to perform a complex discrete search for the
likeliest word sequence. 

Recently, several research groups have begun to study whether a neural network
itself can subsume the functions of the decision tree, and decoding search.
We refer to these approaches as ``all-neural,'' and they are also 
commonly referred to as ``end-to-end.''
Two main approaches have been used, both of which attempt to leverage a 
recurrent neural network's potential ability to ``remember'' information
for a long period of time%
\footnote{The vanishing gradient problem has been sufficiently overcome by
long-short-term memory (LSTM) \cite{hochreiter1997long} and
recurrent neural networks with rectified linear units (ReLU-RNNs) \cite{le2015simple} that many problems
in machine translation and language processing can be handled regardless.}
and then act on it \cite{sutskever2014sequence,bahdanau2014neural,vinyals2015grammar}. 
The first of these approaches uses an RNN trained in the ``connectionist temporal
classification'' (CTC) framework \cite{graves2006connectionist}
to predict letter rather than phonetic output 
\cite{graves2014towards,sak2015fast,maas2015lexicon,hannun2014deep,miao2015eesen}.
At its base, this approach has no decoder at all, with all logic related to 
language modeling and decoding being implicitly done by the RNN. 
The output of a CTC system
can of course be consumed by a subsequent decoding process, but it is not
necessary.  The second approach uses an RNN with an attention mechanism
\cite{bahdanau2016end,lu2016training,chan2016listen}. 
The attention mechanism provides a weighted sum of the hidden 
activations in a learned encoding of the 
input frames as an additional input to the RNN at each time frame. When the
network learns an attention function that happens to be unimodal with a 
peak that moves from left to right monotonically, it is similar to a 
Viterbi alignment.

Attention-based models use a beam search to decode sequentially, symbol by
symbol.

This paper is motivated by the desire to see to what extent the decoding
process of a standard system can be modeled by a neural net itself,
and to gauge the necessity of a pronunciation dictionary and decision tree.
We describe a CTC based system that advances the 
state-of-the-art in all-neural modeling for conversational speech recognition.
We propose a two-stage CTC process, in which we first train a system that
consumes speech features and hypothesizes letter sequences.
Second, we re-use the CTC apparatus to train a system that consumes this noisy letter
sequence and produces a cleaner version.  Additionally, we present a 
careful exploration of the unit-vocabulary, and of the training 
process. We 
find that a symbol inventory that uses special word-initial
characters (capital letters) rather than spaces performs well.
Finally, we explore the addition of both character and word language models. 
We advance the state of the art  at every level from pure all-neural ASR
through to the addition of a word-based decoding process.

The remainder of this paper is organized as follows. Section \ref{sec:prior}
places our work in the context of previous efforts. Section
\ref{sec:model} describe the model we use, and \ref{sec:post} our standard decoding process, and
extensions.  Section \ref{sec:itctc} proposes the novel technique of
iterated CTC. In Section \ref{sec:training} we describe the details of training
the models. Section \ref{sec:experiments} presents experimental results,
followed by conclusions in Section \ref{sec:conc}. 

\section{Relation to Prior Work}
\label{sec:prior}

Letter-based or graphemic systems have been long 
studied \cite{schukat1993automatic,schillo2000grapheme,kanthak2002context,killer2003grapheme}, and are attractive because they
alleviate the need to produce a dictionary of word pronunciations. Past work
was motivated by the need to quickly build systems in new languages
without a dictionary, and kept the 
rest of a standard HMM system, in particular the use of a decision tree.

In contrast to this,
we follow recent work \cite{lu2016training,maas2015lexicon,chan2016listen,hannun2014deep} where a 
neural network learns context-dependence implicitly.

Our approach is most similar to the CTC methods of \cite{miao2015eesen,miao2016empirical,hannun2014deep,maas2015lexicon,sak2015fast}.
In contrast to \cite{miao2015eesen,miao2016empirical,sak2015fast}, we use a ReLU-RNN rather than an LSTM, and find
it to be effective and much faster. In contrast to \cite{hannun2014deep}, 
we use recurrent networks at every level as opposed to deep neural nets (DNNs) in the lower 
levels, and an RNN at the top level only. Also in contrast to \cite{hannun2014deep},
we study performance in the absence of an external language model as well
as with one.

We extend past CTC work by the use of
what we term {\it iterated CTC}, 
first operating on acoustic features, and then  on
letter sequences. This use of CTC on symbolic input is a novel alternative
to encoder-decoder models, and is described in Section \ref{sec:itctc}.
Interestingly, the attention-based approach of \cite{lu2016training} also 
introduces an extra RNN layer with the motivation of modeling symbol/language
level phenomena.

\section{The Model}
\label{sec:model}

We adopt a multi-layer RNN trained with CTC \cite{graves2006connectionist}. Central to the
CTC process is the use of a ``don't care'' or blank symbol, which is 
allowed to optionally occur between regular symbols.

The standard alpha-beta recursions are used to compute the posterior state
occupancy probabilities. 

Let the input consist of $t$ acoustic frames along with
a symbol sequence $S$. Denote an alignment of the $t$ audio frames to the 
sequence $S$ by $\pi$, and the product of the state-level neural 
net probabilities for the alignment as $P(S|\pi)$. Let 
$p_q^t$ be the probability the neural net assigns to symbol $q$ at time $t$, 
i.e., the output after the softmax function. The CTC objective function
is given by 
\[{\cal L} = \sum_{\pi} P(S|\pi) P(\pi) = \sum_{\pi} P(\pi) \prod_t p_{S_{\pi(t)}}^{t}.\]
$P(\pi)$ is determined by
the HMM transition probabilities.
We use a self-loop probability of 0.5 for all symbols.
The probability of transitioning from a non-blank symbol to the blank symbol is 0.25,
and from a non-blank symbol to the next non-blank symbol it is 0.25.
In addition, the probability of transitioning out of the blank symbol to the next non-blank symbol is 0.25.
The key input to CTC is the probability, as determined by the neural network, of a particular symbol $S_t$ at time $t$. 

Consistent with standard notation, denote
the posterior probability of being in state/symbol $q$ at time $t$ by
$\gamma_{q}^{t}$. Note that this is derived from the alpha-beta computations,
and is distinct from the probability $p_q^t$ that the neural network 
assigns to symbol
$q$ at time $t$. The derivative of the CTC objective function with respect to
the activation $a_{q}^{t}$ for output $q$ at time $t$ before the softmax is
\[ \frac{d\cal{L}}{da_{q}^{t}} = \gamma_{q}^{t} - p_{q}^{t} \]
This is the error signal for backpropagation into the RNN.

\section{Interpreting the Output}
\label{sec:post}

\subsection{Raw CTC Output}

After training, the output of the RNN can be directly converted into 
a readable character sequence. There are two problems to solve:
\begin{enumerate} 
\item Where to put spaces between words.
\item How to distinguish instances of repeated characters, 
for example the {\it ll} in {\it hello}, from a sequence of frames each labeled
with the same letter, e.g., the {\it l} in {\it help}. Recall that the CTC
blank symbol is optional, so a sequence of frames labeled by a single letter
cannot be immediately distinguished from multiple occurrences of that letter.
\end{enumerate}
When the symbol inventory includes a space
symbol (distinct from the blank symbol), the first problem is easily solved. 
Past work, e.g. \cite{maas2015lexicon}, solve the second problem with a
search over alternatives, or requiring a blank between letters.
Instead, we propose a new symbol inventory as described below.

\subsection{Symbol Inventory}

Past work \cite{maas2015lexicon,hannun2014deep,lu2016training} has explicitly modeled the spaces
between words in the acoustic model, e.g., with a special space symbol ``\_''
distinct from the CTC blank symbol. Since words are frequently run together,
we propose an alternative representation where 
word-initial characters are considered distinct from non-initial characters.
A convenient representation of this is to use capital letters in the
word-initial position. Since the forward-backward computation in
CTC requires that the input sequence be longer
than the output sequence, this also increases the set of utterances that
can be aligned. 
This is also consistent with speech recognition systems that use 
position-dependent phonetic variants. 
To identify repeated letters, we use 
special double-letter units to represent repeated
characters like {\it ll}. Finally, to improve the readability of the
output without any further processing, 
we attach apostrophes to 
the following letters, creating units 
like the ``'d'' in ``we'd.'' Altogether the unit inventory size is 79.
\vspace*{0.1in}

As an example of this, the sentence 
\begin{center}
``{\em yes he has one}'' 
\end{center}
would be rendered for training as  
\begin{center}
``{\em YesHeHasOne}''.
\end{center}
With this encoding, a readable decoding can then be produced very simply:
\begin{enumerate}
\item Select the most likely symbol at each frame
\item Discard all occurrences of the ``don't care'' symbol
\item Compress consecutive occurrences of the same letter into one occurrence 
\item Add a space in front of each capitalized letter and show the output
\end{enumerate}

\subsection{Character Beam Search}
Previous work \cite{hannun2014deep,maas2015lexicon,chan2016listen} has used a
character-level language model to improve the output of a neural system. This implements
the classical decoding paradigm of Eqn. \ref{eq:dcd}

 at the character level, with the character language model providing
$P(w)$ or in this case $P(c)$. The neural network provides $P(c|a)$, and 
beam search is used to find the likeliest character sequence. We 
present results for this approach in Section \ref{ssec:mr}.

\subsection{Word Beam Search}

To provide a complete set of results comparable to \cite{hannun2014deep},
we have also used a word-based decoder that uses a graphemic dictionary,
and uses the frame-level likelihoods in the standard way.
The decoder is the dynamic decoder as described in \cite{mendis2016parallelizing}.
We used the CUED-RNNLM toolkit \cite{chen2016cued}
to train two forward- and two backward-running RNN language models.
These are interpolated with a standard 4-gram model and used to rescore N-best lists produced 
by the N-gram decoder.
Details can be found in a companion paper \cite{ms-swb-icassp2017}.

\section{Iterated CTC (\CTCTWO)}
\label{sec:itctc}

The output described in the previous section is of course noisy. For example,
one of the Switchboard utterances is\
{\it ``no white collar crime does not exactly fall into it''},\
but the raw network output is\
{\it ``and now whi coler crime doen exsitally fall into it''}.\
The classical approach to improving this is the incorporation of a lexicon
of legal word units and a language model, as 
described in  Section \ref{sec:post}.

To improve the output with a purely neural network based approach, we 
propose using iterated CTC.
Specifically, the noisy character sequences from the initial raw CTC output is represented by
one-hot feature vectors analogous to the acoustic feature vectors, and 
the RNN/CTC training process is repeated. The result is a model that 
transforms a noisy symbol sequence into a less noisy sequence. 
We have found that this process, while producing less dramatic improvements
than the incorporation of a full fledged decoder, consistently improves
the output, while staying in the all-neural paradigm.
This is the case even when the original network is optimally deep.

\section{Training Process}
\label{sec:training}

Our models are trained using stochastic gradient descent with momentum
and L2 regularization.
For all but our largest networks, 
minibatches of $32$ utterances are processed at once, resulting in updates
after several thousand speech frames. For networks of width 1024 and depth 
7 or greater, we have found it necessary to process 64 utterances
simultaneously to achieve accurate gradient estimates and stable
convergence. We use frame-skipping \cite{sak2015fast},
where we stack three consecutive frames into a single vector to produce an
input sequence one-third as long and three times as wide as the original input.
When we train on the 300-hour Switchboard set, we use 
a per-frame learning rate of 0.5, and decrease it by a factor of
4 if 3 iterations over the data fail to produce an improvement on 
development data.
We randomly held out about 10 hours of training data for use as development data.
When the Fisher data is added, resulting in about 2000 hours
of training data, the rate is reduced if a single iteration passes without
increasing dev set likelihood.

We implement the model with direct calls to the CUDNN v5.0 RNN library.
Training with 40-dimensional input feature vectors (prior to skipping), a
512-dimension bi-directional ReLU-RNN with three hidden layers is about 0.0025
times real time, i.e., 400 times faster than real time on a Razer laptop with
a NVIDIA 970M GPU.

\begin{table}[t]
\begin{center}
\caption{Word error rate (\%) on NIST RT-02 Switchboard-1 test set as a function of symbol inventory, for a 512-wide 5-deep network}
\vspace*{0.1in}
\label{tab:syms}
\begin{tabular}{|c|c|c|} \hline
Explicit spaces & Capital letters & Initial+Final letters \\ \hline \hline
38.1 & 36.2 & 36.2 \\ \hline
\end{tabular}
\vspace*{-0.1in}
\end{center}
\end{table}

\begin{table}[t]
\begin{center}
\caption{Word error rate (\%) on the RT-02 test set as a function of hidden layer size, for 5-layer networks}
\vspace*{0.1in}
\label{tab:width}
\begin{tabular}{|c|c|} \hline
Hidden Dimension & WER \\ \hline \hline
512 & 36.2  \\ \hline
1024 & 31.9  \\ \hline
2048 & 30.4 \\ \hline
\end{tabular}
\vspace*{-0.1in}
\end{center}
\end{table}

\subsection{Stabilization Methods}

In initial experiments, we found it useful to introduce several
stabilization techniques. Most importantly, we use gradient clipping to 
prevent ``exploding gradients'' during the RNN training. The magnitude of the
gradients are clipped at 1 prior to the momentum update. 
Secondarily, we have noticed that when rare units are present (e.g., the
``ii'' in ``Hawaii,'' the training process tends to push their probabilities
close to 0 between occurrences, which leads to poor performance and
sometimes instability when
the unit is eventually seen. To avoid this, we interpolate the gradient
with a small gradient tending towards the uniform distribution. This is 
implemented by interpolating the $\gamma$ values from the alpha-beta computation
with a uniform distribution.
We reserve 1\% of the total probability mass for this uniform distribution.
Finally, we have also found it important to compute the gradient over a large 
number of utterances (32 or 64) before doing a parameter update.

\subsection{Model Initialization}
Weight matrices are initialized with small random weights 
uniformly distributed and inversely 
proportional to the square root of the fan-in, with one exception. In the
output layer, which maps from the RNN hidden dimension (typically 1024)
down to the
size of the symbol inventory (79 in our case), 
we assign explicit responsibility for
each output symbol to a specific RNN activation. This is done by 
using an identity matrix for the first 79 dimensions, and zeros elsewhere. 
In the case of bidirectional networks, this is done symmetrically so 
both forward and backward portions of the network contribute equally. 
While we have not performed an exhaustive evaluation of this scheme, in
initial experiments we observed consistent small gains. 
Concurrent with this work, a similar scheme was 
very recently proposed in \cite{kurata2016improved} for
standard neural net systems.

\subsection{Polishing with In-domain Data}

Our training process begins and ends with a focus on the in-domain
300 hour switchboard-only dataset. We start by training on the 300 hour
set mainly for convenience in showing results with both the 300 hour (Switchboard) and 
2000 hour (Switchboard + Fisher) data-sets.  The 2000 hour models presented
here were initialized with the output of 300 hour training.
While training from scratch with 2000 hours of data works about as well,
we have found it consistently useful to finish all training runs by
executing a few more iterations of training on the
in-domain Switchboard-only data. We do this starting from a very low
learning rate (one-tenth the normal rate), and most of the gain is observed
in the first iteration of training.

\section{Experiments}
\label{sec:experiments}

\begin{table}[t]
\begin{center}
\caption{Word error rate (\%) as a function of the number of layers and dimensions
on the RT-02 test set}
\vspace*{0.1in}
\label{tab:layers}
\begin{tabular}{|c|c|c|} \hline
Number of layers & 512 width & 1024 width \\ \hline \hline
3 & 48.5 &  38.1 \\ \hline
4 & 38.3 &  34.6 \\ \hline
5 & 36.2 & 31.9 \\ \hline
6 & 34.5 & 30.3 \\ \hline
7 & 32.5 & 30.6 \\ \hline
8 & 31.0 & 31.3 \\ \hline
9 & 31.4 & 29.4 \\ \hline
10 & 31.5 &  29.4 \\ \hline
\end{tabular}
\end{center}
\vspace*{-0.1in}
\end{table}

\begin{table}[t]
\begin{center}
\caption{Word error rate (\%) as a function of post-processing for the RT-02 test set, using
a 9-layer 1024-wide network}
\vspace*{0.1in}
\label{tab:post}
\begin{tabular}{|c|c|c|c|} \hline
None & Iterated CTC & Character-Beam & Word-Beam \\ \hline \hline
29.4 & 27.9 & 23.3 & 19.2  \\ \hline
\end{tabular}
\end{center}
\vspace*{-0.1in}
\end{table}

\subsection{Corpus}

For comparability with \cite{maas2015lexicon,hannun2014deep,chan2016listen,lu2016training}, we
present results on the NIST 2000 conversational telephone speech (CTS) evaluation set.
Model selection uses the RT-02 CTS evaluation set for development.
The input features are 40-dimensional log-Mel-filterbank energies, 
extracted every
10 milli\-seconds. The feature vectors are normalized to zero mean on 
a per-utterance level. Since the logarithmic compression already limits
the dynamic range to reasonable levels, we do not perform variance 
normalization.

\subsection{Network Architecture and Symbol Inventory}

For computational efficiency, we restricted ourselves to the models
directly supported by the CUDNN v5.0 library. This encompasses 
multi-layer uni- and bidirectional RNNs. Support is provided for 
LSTMs, Gated Recurrent Units,
standard sigmoid RNNs, and ReLU-RNNs.  In initial experiments, we
found that ReLU-RNNs are as good as LSTMs for this task, and many times
faster. Therefore we use them exclusively in the experiments. We further
focus on bi-directional networks for improved performance.

\begin{table}[t]
\begin{center}
\caption{Comparative performance: word error rates (\%) on the NIST 2000 Switchboard and CallHome test sets (models trained on 300 hours of Switchboard data only). All systems are use graphemic targets.}
\vspace*{0.1in}
\label{tab:comps}
\begin{tabular}{|l|l|l|l|l|l|} \hline
Reference & Lexicon & LM & CH & SW \\ \hline \hline
\cite{maas2015lexicon} & N & N & 56.1  & 38.0 \\ \hline
\cite{lu2016training} & N & N & 48.2 &  27.3 \\ \hline
Current & N & N & 38.8 & 25.9 \\ \hline
Current+\CTCTWO & N & N & {\bf 37.1} & {\bf 24.7} \\ \hline \hline
\cite{maas2015lexicon} & N & Char NG & 43.8 & 27.8  \\ \hline
\cite{maas2015lexicon} & N & Char RNN & 40.2 & 21.4 \\ \hline
Current & N & Char NG & {\bf 32.1 } & {\bf 19.8} \\ \hline \hline
\cite{lu2016training} & Y & Word NG & 46.0 & 25.8 \\ \hline
\cite{hannun2014deep} & Y & Word NG & 31.8 & 20.0  \\ \hline

Current & Y & Word NG &  26.3 & 15.1 \\ \hline 
Current & Y & Word RNN & {\bf 25.3} & {\bf 14.0} \\ \hline 
\end{tabular}
\end{center}
\vspace*{-0.1in}
\end{table}

Table \ref{tab:syms} shows the effect of our choice of symbol
inventory. We see that the use of special word-initial characters 
improves performance over the use of explicit blanks. Redundantly 
modeling word-final characters does not provide a further improvement.
For ease of interpretability, all models use explicit double-character 
symbols.

In Table \ref{tab:width}, we show the effect of network width, keeping the
depth constant at 5. While the widest network 
is the best, for computational reasons we restrict our further
experiments to networks of width 1024 or less.

In Table \ref{tab:layers}, we present the  effect of network depth,
for hidden layer sizes of 512 and 1024. Note that since
we use a bidirectional network, the total number of activations
in a layer is double this. We find that 
relatively deep networks perform well.
Based on these results, the remainder of the experiments use a 
1024 width 9 layer bidirectional network.
Including weight matrices and biases, the total number of parameters
in the 9 layer 1024 wise network is about 53 million parameters.

\subsection{Iterated CTC and Beam Search}

We evaluate the post-processing methods in Table \ref{tab:post}.
Clearly, the RNN is not yet learning all the logic of a beam-search decoder,
and the effectiveness of a character-based beam search is midway between
using the raw output, and a full word-based search. We see that iterated
CTC can produce a significant improvement, though not as much as a 
complete beam search, while remaining in the all-neural framework.
When character or word N-grams are used, the utility decreases.
An examination of the iterated CTC errors indicates that it mostly
reduces the substitution rates, as the global shifts created by 
insertions and deletions seem difficult for the RNN to compensate.

\subsection{Comparison to Previously Published Results}
\label{ssec:mr}

We summarize our results on the NIST 2000 CTS test set and compare them with 
past work in Tables~\ref{tab:comps} and~\ref{tab:comps2k}. The
systems are categorized according to 
whether they use a lexicon to enforce the output of legal words, and in
their use of a language model. We see an improvement over 
previous results with our RNN based system. 
In \cite{miao2016empirical}, a LSTM-CTC system
using {\it phonemic} rather than graphemic targets is presented, and achieves
an error rate of 15\% on the Switchboard portion of eval 2000; that
system still uses a phonetic dictionary.
Note that the authors in \cite{miao2016empirical} did not report an error rate on Switchboard tasks
with characters as target.
The only reported number for a CTC-based system with character output is from \cite{hannun2014deep},
where the error rate is 20.0\%.
This is much higher than the 15\% error rate we obtained in this work under the same evaluation conditions.  
Compared to a standard system, such as \cite{povey2016purely}, which achieves
9.6\% and 13\% on Switchboard and CallHome respectively
with conventional 300-hour training,
we see that current neural-only systems cannot yet mimic all the 
logic in a conventional system.  However, our ReLU-RNN system does set a 
new state of the art for an all-neural system, and we see that performance 
of such systems is rapidly improving.

\begin{table}[t]
\begin{center}
\caption{Comparative performance: word error rate (\%) on the NIST 2000  Switchboard and CallHome test sets 
(models trained on 2000 hours of combined Fisher \& Switchboard data)}
\vspace*{0.1in}
\label{tab:comps2k}
\begin{tabular}{|l|l|l|l|l|l|} \hline
Reference & Lexicon & LM & CH & SW \\ \hline \hline
Current  & N & N & 26.4  & 17.2 \\ \hline \hline
Current  & N & Char NG & 21.8  & 13.8 \\ \hline \hline
\cite{hannun2014deep} (ensemble) & Y & Word NG & 19.3  & 12.6 \\ \hline
Current & Y & Word NG & 18.7 & 11.3 \\ \hline 
Current & Y & Word RNN & {\bf 17.7}  & {\bf 10.2} \\ \hline 
\end{tabular}
\end{center}
\vspace*{-0.1in}
\end{table}

\section{Conclusions}
\label{sec:conc}

We advance the state of the art with an all-neural 
speech recognizer, principally by employing a novel symbol encoding, and
optimized training process.
We further present an iterated CTC approach for use without any 
decoding process.
In this framework, a network first maps from audio to symbols,
followed by a second symbol-to-symbol mapping network.

Both using raw network output and search-based post-processing, we 
systematically improve
on previously published results in the end-to-end neural speech recognition paradigm.

\footnotesize
\bibliographystyle{ieee-shortnames}
\bibliography{strings,refs}

\begin{thebibliography}{10}

\bibitem{dahl2011large}
G.~E. Dahl, D.~Yu, L.~Deng, and A.~Acero,
\newblock ``Large vocabulary continuous speech recognition with
  context-dependent {DBN-HMMs}'',
\newblock {\em in} {\em ICASSP}, pp. 4688--4691. IEEE, 2011.

\bibitem{mohamed2009deep}
A.-r. Mohamed, G.~Dahl, and G.~Hinton,
\newblock ``Deep belief networks for phone recognition'',
\newblock {\em in} {\em NIPS Workshop on Deep Learning for Speech Recognition
  and Related Applications}, p.~39, 2009.

\bibitem{seide2011conversational}
F.~Seide, G.~Li, and D.~Yu,
\newblock ``Conversational speech transcription using context-dependent deep
  neural networks'',
\newblock {\em in} {\em Interspeech}, pp. 437--440, 2011.

\bibitem{sak2014long}
H.~Sak, A.~W. Senior, and F.~Beaufays,
\newblock ``Long short-term memory recurrent neural network architectures for
  large scale acoustic modeling'',
\newblock {\em in} {\em Interspeech}, pp. 338--342, 2014.

\bibitem{hinton2012deep}
G.~Hinton, L.~Deng, D.~Yu, G.~E. Dahl, A.-r. Mohamed, N.~Jaitly, A.~Senior,
  V.~Vanhoucke, P.~Nguyen, T.~N. Sainath, et~al.,
\newblock ``Deep neural networks for acoustic modeling in speech recognition:
  The shared views of four research groups'',
\newblock {\em IEEE Signal Processing Magazine}, vol. 29, pp. 82--97, 2012.

\bibitem{bourlard2012connectionist}
H.~A. Bourlard and N.~Morgan,
\newblock {\em Connectionist speech recognition: a hybrid approach},
\newblock Springer Science \& Business Media, 1993.

\bibitem{robinson1991recurrent}
T.~Robinson and F.~Fallside,
\newblock ``A recurrent error propagation network speech recognition system'',
\newblock {\em Computer Speech \& Language}, vol. 5, pp. 259--274, 1991.

\bibitem{hermansky2000tandem}
H.~Hermansky, D.~P. Ellis, and S.~Sharma,
\newblock ``Tandem connectionist feature extraction for conventional {HMM}
  systems'',
\newblock {\em in} {\em ICASSP 2000}, vol.~3, pp. 1635--1638. IEEE, 2000.

\bibitem{young1994tree}
S.~J. Young, J.~J. Odell, and P.~C. Woodland,
\newblock ``Tree-based state tying for high accuracy acoustic modelling'',
\newblock {\em in} {\em Proceedings of the workshop on Human Language
  Technology}, pp. 307--312. Association for Computational Linguistics, 1994.

\bibitem{hochreiter1997long}
S.~Hochreiter and J.~Schmidhuber,
\newblock ``Long short-term memory'',
\newblock {\em Neural computation}, vol. 9, pp. 1735--1780, 1997.

\bibitem{le2015simple}
Q.~V. Le, N.~Jaitly, and G.~E. Hinton,
\newblock ``A simple way to initialize recurrent networks of rectified linear
  units'',
\newblock arXiv preprint arXiv:1504.00941, 2015.

\bibitem{sutskever2014sequence}
I.~Sutskever, O.~Vinyals, and Q.~V. Le,
\newblock ``Sequence to sequence learning with neural networks'',
\newblock {\em in} {\em Advances in neural information processing systems}, pp.
  3104--3112, 2014.

\bibitem{bahdanau2014neural}
D.~Bahdanau, K.~Cho, and Y.~Bengio,
\newblock ``Neural machine translation by jointly learning to align and
  translate'',
\newblock arXiv preprint arXiv:1409.0473, 2014.

\bibitem{vinyals2015grammar}
O.~Vinyals, {\L}.~Kaiser, T.~Koo, S.~Petrov, I.~Sutskever, and G.~Hinton,
\newblock ``Grammar as a foreign language'',
\newblock {\em in} {\em Advances in Neural Information Processing Systems}, pp.
  2773--2781, 2015.

\bibitem{graves2006connectionist}
A.~Graves, S.~Fern{\'a}ndez, F.~Gomez, and J.~Schmidhuber,
\newblock ``Connectionist temporal classification: labelling unsegmented
  sequence data with recurrent neural networks'',
\newblock {\em in} {\em Proc.\ of the 23rd Intl.\ Conf.\ on Machine learning},
  pp. 369--376. ACM, 2006.

\bibitem{graves2014towards}
A.~Graves and N.~Jaitly,
\newblock ``Towards end-to-end speech recognition with recurrent neural
  networks'',
\newblock {\em in} {\em ICML}, vol.~14, pp. 1764--1772, 2014.

\bibitem{sak2015fast}
H.~Sak, A.~Senior, K.~Rao, and F.~Beaufays,
\newblock ``Fast and accurate recurrent neural network acoustic models for
  speech recognition'',
\newblock {\em in} {\em Interspeech}, pp. 1468--1472, 2015.

\bibitem{maas2015lexicon}
A.~L. Maas, Z.~Xie, D.~Jurafsky, and A.~Y. Ng,
\newblock ``Lexicon-free conversational speech recognition with neural
  networks'',
\newblock {\em in} {\em Proc. NAACL}, pp. 345--354, 2015.

\bibitem{hannun2014deep}
A.~Hannun, C.~Case, J.~Casper, B.~Catanzaro, G.~Diamos, E.~Elsen, R.~Prenger,
  S.~Satheesh, S.~Sengupta, A.~Coates, et~al.,
\newblock ``Deep speech: Scaling up end-to-end speech recognition'',
\newblock arXiv preprint arXiv:1412.5567, 2014.

\bibitem{miao2015eesen}
Y.~Miao, M.~Gowayyed, and F.~Metze,
\newblock ``Eesen: End-to-end speech recognition using deep rnn models and
  wfst-based decoding'',
\newblock {\em in} {\em IEEE ASRU Workshop}, pp. 167--174. IEEE, 2015.

\bibitem{bahdanau2016end}
D.~Bahdanau, J.~Chorowski, D.~Serdyuk, Y.~Bengio, et~al.,
\newblock ``End-to-end attention-based large vocabulary speech recognition'',
\newblock {\em in} {\em ICASSP}, pp. 4945--4949. IEEE, 2016.

\bibitem{lu2016training}
L.~Lu, X.~Zhang, and S.~Renals,
\newblock ``On training the recurrent neural network encoder-decoder for large
  vocabulary end-to-end speech recognition'',
\newblock {\em in} {\em ICASSP}, pp. 5060--5064. IEEE, 2016.

\bibitem{chan2016listen}
W.~Chan, N.~Jaitly, Q.~Le, and O.~Vinyals,
\newblock ``Listen, attend and spell: A neural network for large vocabulary
  conversational speech recognition'',
\newblock {\em in} {\em ICASSP}, pp. 4960--4964. IEEE, 2016.

\bibitem{schukat1993automatic}
E.~G. Schukat-Talamazzini, H.~Niemann, W.~Eckert, T.~Kuhn, and S.~Rieck,
\newblock ``Automatic speech recognition without phonemes'',
\newblock {\em in} {\em Eurospeech}, pp. 129--132, 1993.

\bibitem{schillo2000grapheme}
C.~Schillo, G.~A. Fink, and F.~Kummert,
\newblock ``Grapheme based speech recognition for large vocabularies'',
\newblock {\em in} {\em Interspeech}, pp. 584--587, 2000.

\bibitem{kanthak2002context}
S.~Kanthak and H.~Ney,
\newblock ``Context-dependent acoustic modeling using graphemes for large
  vocabulary speech recognition'',
\newblock {\em in} {\em ICASSP}, vol.~2, pp. 845--848, 2002.

\bibitem{killer2003grapheme}
M.~Killer, S.~St{\"u}ker, and T.~Schultz,
\newblock ``Grapheme based speech recognition'',
\newblock {\em in} {\em Interspeech}, pp. 3141--3144, 2003.

\bibitem{miao2016empirical}
Y.~Miao, M.~Gowayyed, X.~Na, T.~Ko, F.~Metze, and A.~Waibel,
\newblock ``An empirical exploration of ctc acoustic models'',
\newblock {\em in} {\em ICASSP}, pp. 2623--2627. IEEE, 2016.

\bibitem{mendis2016parallelizing}
C.~Mendis, J.~Droppo, S.~Maleki, M.~Musuvathi, T.~Mytkowicz, and G.~Zweig,
\newblock ``Parallelizing wfst speech decoders'',
\newblock {\em in} {\em ICASSP}, pp. 5325--5329. IEEE, 2016.

\bibitem{chen2016cued}
X.~Chen, X.~Liu, Y.~Qian, M.~Gales, and P.~Woodland,
\newblock ``{CUED-RNNLM}: An open-source toolkit for efficient training and
  evaluation of recurrent neural network language models'',
\newblock {\em in} {\em ICASSP}, pp. 6000--6004. IEEE, 2016.

\bibitem{ms-swb-icassp2017}
W.~Xiong, J.~Droppo, X.~Huang, F.~Seide, M.~Seltzer, A.~Stolcke, D.~Yu, and
  G.~Zweig,
\newblock ``The {Microsoft} 2016 conversational speech recognition system'',
\newblock {\em in} {\em ICASSP}, 2017.

\bibitem{kurata2016improved}
G.~Kurata and B.~Kingsbury,
\newblock ``Improved neural network initialization by grouping
  context-dependent targets for acoustic modeling'',
\newblock {\em in} {\em Interspeech}, pp. 27--31, 2016.

\bibitem{povey2016purely}
D.~Povey, V.~Peddinti, D.~Galvez, P.~Ghahrmani, V.~Manohar, X.~Na, Y.~Wang, and
  S.~Khudanpur,
\newblock ``Purely sequence-trained neural networks for {ASR} based on
  lattice-free {MMI}'',
\newblock {\em in} {\em Interspeech}, pp. 2751--2755, 2016.

\end{thebibliography}

\end{document}